\def\year{2022}\relax
\documentclass[letterpaper]{article} 
\usepackage{aaai22}  
\usepackage{times}  
\usepackage{helvet}  
\usepackage{courier}  
\usepackage[hyphens]{url}  
\usepackage{graphicx} 
\usepackage{natbib}  
\usepackage{caption} 
\DeclareCaptionStyle{ruled}{labelfont=normalfont,labelsep=colon,strut=off} 
\usepackage{amsmath}
\usepackage{amsfonts}
\usepackage{graphicx}
\usepackage{multirow}
\usepackage{lipsum}
\usepackage{bbding}
\usepackage{placeins}

\usepackage{booktabs}  %
\usepackage{multirow}

\usepackage{algorithm}
\usepackage{algorithmic}

\usepackage[ruled,algo2e]{algorithm2e}

\usepackage{newfloat}
\usepackage{listings}
\floatstyle{ruled}
\newfloat{listing}{tb}{lst}{}
\floatname{listing}{Listing}
\usepackage{algorithm}
\usepackage{algorithmic}

\title{Hybrid Curriculum Learning for Emotion Recognition in Conversation}
\author{
Lin Yang\equalcontrib, Yi Shen\equalcontrib, Yue Mao, Longjun Cai
}
\affiliations{
 Alibaba Group, Beijing, China\\
  \{yl176562, sy133447, maoyue.my, longjun.clj\}@alibaba-inc.com
}

\usepackage{bibentry}

\begin{document}

\maketitle

\begin{abstract}

Emotion recognition in conversation (ERC) aims to detect the emotion label for each utterance. 
Motivated by recent studies which have proven that feeding training examples in a meaningful order rather than considering them randomly can boost the performance of models, we propose an ERC-oriented hybrid curriculum learning framework. Our framework consists of two curricula: (1) conversation-level curriculum (CC); and (2) utterance-level curriculum (UC). In CC, we construct a difficulty measurer based on ``emotion shift'' frequency within a conversation, then the conversations are scheduled in an ``easy to hard" schema according to the difficulty score returned by the difficulty measurer. For UC, it is implemented from an emotion-similarity perspective, which progressively strengthens the model’s ability in identifying the confusing emotions. With the proposed model-agnostic hybrid curriculum learning strategy, we observe significant performance boosts over a wide range of existing ERC models and we are able to achieve new state-of-the-art results on four public ERC datasets.

\end{abstract}

\section{Introduction}

Emotion recognition in conversation (ERC) has attracted numerous interests from the NLP community  in recent years due to its potential applications in many areas, such as opinion mining in social media \cite{chatterjee2019semeval}, dialogue generation \cite{huang2018automatic} and fake news detection \cite{guo2019dean}. The objective of ERC is to detect emotions expressed by the speakers in each utterance of the conversation. Previous works on ERC usually solve this problem with two steps. At the first step, each utterance is encoded separately into an utterance-level representation, which will be used as the input for sequence-based models \citep{majumder2019dialoguernn,hazarika2018icon,jiao2019higru} or graph-based models \citep{ghosal2019dialoguegcn,ishiwatari2020relation} during the second step. Despite their success, previous works still have a lot of room for improvement \cite{poria2019emotion}.

Curriculum learning (CL) \cite{bengio2009curriculum} is a training strategy which imitates the meaningful learning order in human curricula. The core idea of CL is to train the machine learning model with easier data subsets at first, and then gradually increase the difficulty level of data until the whole training dataset. As an easy-to-use plug-in, the CL strategy has demonstrated its power in improving the overall performance of various models in a wide range of scenarios \cite{wang2020survey}. Inspired by the success of CL in other NLP tasks \cite{zhou2020uncertainty,liu2018curriculum,su2020dialogue}, in this paper, we make effort to leverage the spirit of CL to improve the traditional ERC methods. Due to the hierarchical structure of the ERC datasets, we need to construct the curricula from two granularities: one curriculum sorts the conversations in the dataset from easy to hard, and the other sorts the utterances in each conversation from easy to hard. 

The question arises how to measure the difficulty of conversations and utterances.  Previous studies \cite{majumder2019dialoguernn,shen2021dialogxl} have reported that most ERC methods mainly suffer from two issues:
1) \emph{``emotion shift'' problem}. It refers to that these methods cannot efficiently handle scenarios in which emotions of two consecutive utterances are different \cite{ghosal-etal-2021-exploring}. 2) \emph{``confusing label'' problem}. Previous methods \cite{ghosal2019dialoguegcn,shen2021directed} usually fail to distinguish between similar emotions very well. This is due to the subtle semantic difference between certain emotion labels such as \emph{happy} and \emph{exciting}. These two phenomena provide us the key to quantify the difficulty of conversations and utterances in ERC.

In this paper, we tailor-design a hybrid curriculum learning (HCL) framework for the ERC task. HCL framework consists of two complementary curriculum strategies, conversation-level curriculum (CC) and utterance-level curriculum (UC). 
In CC, we construct a difficulty measurer based on ``emotion shift" frequency within a conversation, then the conversations with lower difficulty are presented to the model before harder ones. This way, the model gradually increases its ability to tackle the ``emotion shift" problem. 

While in UC, since ERC requires reasoning over multiple utterances in the conversation, we cannot directly schedule the utterances asynchronously in the “easy to hard” scheme. As a result, we design an emotion-similarity based curriculum (ESC) to implement utterance-level curriculum learning. Specifically, inspired by the \emph{``confusing label'' problem} mentioned above, we believe that in a conversation, those utterances with confusing labels are more difficult than others.  Therefore, we make the model focus on the utterances with easily recognizable emotion labels in the early stage, then progressively strengthened the model’s capability of identifying the confusing emotions.

More specifically, based on previous studies \cite{plutchik1982psychoevolutionary,mikels2005emotional} on psychology,  we employ the degree of intersection angle between different emotion labels in Valence-Arousal 2D emotion space \cite{guo2019dean,yang2021circular} to measure the similarity between emotion labels. 

During ESC, instead of one-hot encoding, the target is a probability distribution over all possible emotion labels. The probability of each label is determined by the similarity between current label and the gold label. In other words, besides its true emotion label, each utterance can also belong to similar emotions to a lesser extent.  In the beginning of the training process, the targets of utterances with emotions \emph{happy} and \emph{excited} should almost be the same, but always be very different from \emph{sad}. Then the label representation gradually shifted to the one-hot encoding during the following training process. This way, small mistakes are corrected less than serious mistakes at the beginning, it could also be considered as an easy-to-hard curriculum in which broad concepts are taught before those tiny differences are emphasized. 

Our hybrid curriculum learning framework is model-agnostic. We evaluate our approach on five representative ERC models. Results on four benchmark datasets demonstrate that the proposed hybrid curriculum learning framework leads to significant performance improvements.

In summary, our main contributions are as follows:

\begin{itemize}
	\item We propose a hybrid curriculum learning framework to tackle the task of ERC. At conversation-level curriculum, we utilize an emotion-shift frequency to measure the difficulty of each conversation.
	\item We propose emotion-similarity based curriculum learning to achieve utterance-level curriculum learning. It implements the basic idea that at early stage of training it is less important to distinguish between similar emotions compared to separating very different emotions. 
	\item We conduct experiments  on four ERC benchmark datasets. Empirical results show that our proposed hybrid curriculum learning framework can effectively improve the overall performance of various ERC models,  including the state-of-the-art. 	
\end{itemize}

\section{Related Work}

\subsection{Emotion Recognition in Conversation}

Emotion recognition in conversations (ERC) has been widely studied due to its potential application prospect. The key point of ERC is how to effectively model the context of each utterance and corresponding speaker. Existing works generally resort to deep learning methods to capture contextual characteristics, which can be divided into sequence-based and graph-based methods. Another direction is to improve the performance of existing models by incorporating various external knowledge, which we classified as knowledge-based methods. 

\textbf{Sequence-based Methods} Many previous works consider contextual information as utterance sequences. ICON \cite{hazarika2018icon} and CMN \cite{hazarika2018conversational} both utilize gated recurrent unit (GRU) to model the utterance sequences.  DialogueRNN \cite{majumder2019dialoguernn} employs a GRU to capture the global context which is updated by the speaker state GRUs. \citet{jiao2019higru} propose a hierarchical neural network model that comprises two GRUs for the modelling of tokens and utterances respectively. \citet{hu2021dialoguecrn} introduce multi-turn reasoning modules on Bi-directional LSTM to model the ERC task from a cognitive perspective.

\textbf{Graph-based Methods} In this category, some existing works \cite{ghosal2019dialoguegcn,ishiwatari2020relation,zhang2019modeling} utilize various graph neural networks to capture multiple dependencies in the conversation. DialogXL \cite{shen2021dialogxl} modifies the memory block in XLNet \cite{yang2019xlnet} to store historical context and leverages the self-attention mechanism in XLNet to deal with the multi-turn multi-party structure in conversation. \citet{shen2021directed} design a directed acyclic graph (DAG) to model the intrinsic structure within a conversation, which achieves the state-of-the-art performance without considering the introduction of external knowledge.

\textbf{Knowledge-based Methods} KET \cite{zhong2019knowledge}  employs hierarchical transformers with concept representations extracted from the ConceptNet \cite{speer2017conceptnet} for emotion detection, which is the first ERC model integrates common-sense knowledge. COSMIC \cite{ghosal2020cosmic} adopts a network structure very close to DialogRNN and adds external commonsense knowledge from ATOMIC \cite{sap2019atomic} to improve its performance. TODKAT \cite{zhu2021topic} leverages an encoder-decoder architecture which incorporates topic representation with commonsense knowledge from ATOMIC for ERC.

\subsection{Curriculum Learning}

Starting from the work by \citet{bengio2009curriculum}, a variety of curriculum learning approaches \citep{wang2020survey, soviany2021curriculum} has been studied. In the field of NLP, curriculum learning has been used for various tasks such as neural machine translation \citep{zhou2020uncertainty,liu2020norm},  relation extraction \cite{huang2019self} and  natural answer generation \cite{liu2018curriculum}. To the best of our knowledge,  we leverage  curriculum learning for the first time in the ERC task.

\section{Proposed Framework}

\subsection{Task Definition}

In ERC, a conversation \emph{C} contains a  sequence of textual utterances  $  \{ u_1, u_2...,u_n \}$, where $n$ denotes the number of utterances. Each utterance $u_i = \{ w_{i,1}, w_{i,2} ..., w_{i,t(u_{i})}\} $ consists of $t(u_{i})$ tokens, where $t(u_{i})$ is the length of $u_i$. There are $m$ participants  $ P =  \{ p_1, p_2...,p_m \} ( m \geq 2)$ in  \emph{C}.  Each utterance $u_i$ is uttered by one participant in $P$ . Given a pre-defined emotion label set $E= \{y_1, y_2, ... , y_r\}$,  the objective of the ERC task is to predict the emotion label of each utterance in \emph{C} with the information provided above.

\begin{figure*}[htbp]  	
	\centering  
	\includegraphics[scale=0.7]{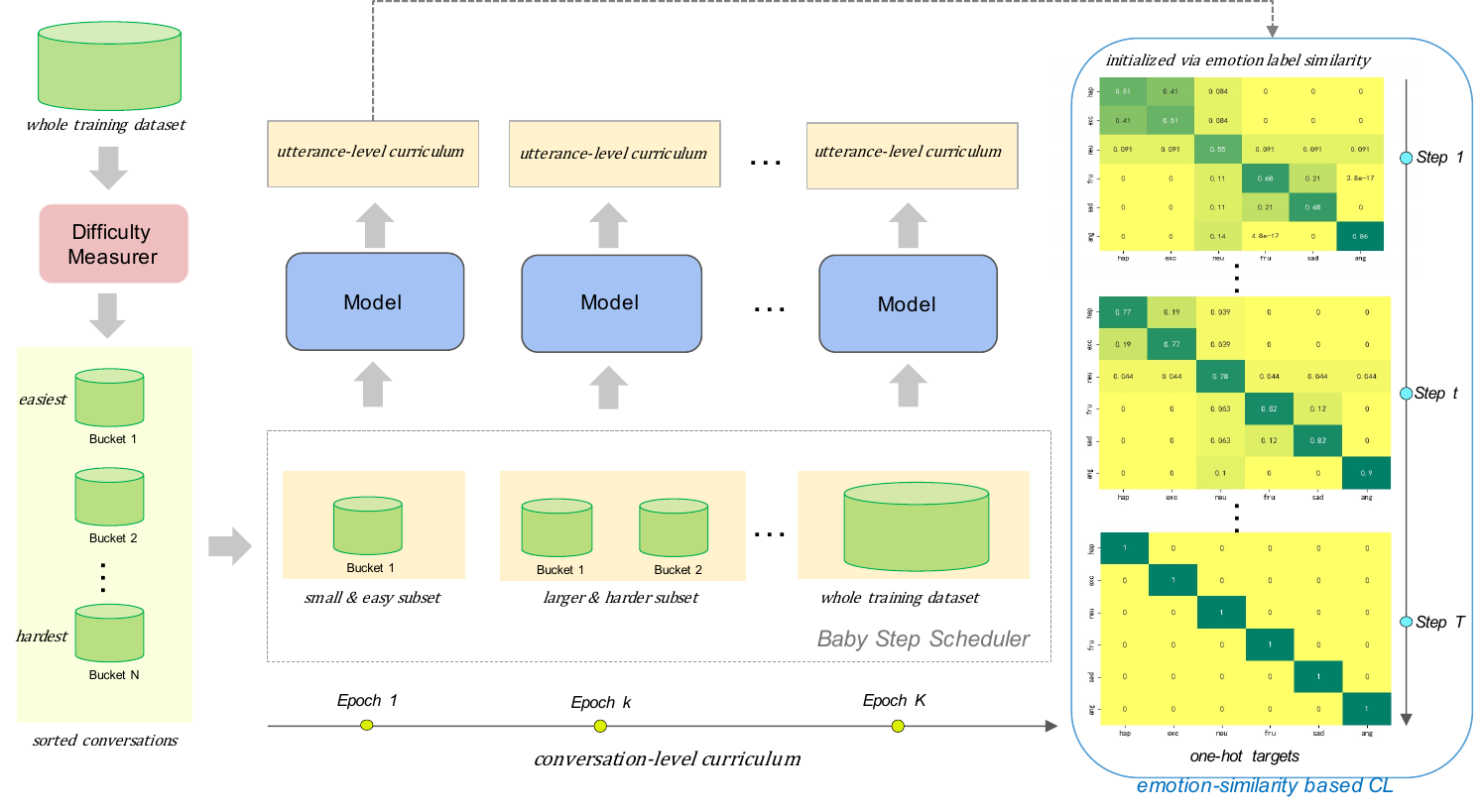}
	\caption{The proposed hybrid curriculum learning (HCL) framework for ERC.}  
	\label{fig:hcl}  	
\end{figure*}

\subsection{Overview}

In curriculum learning, a typical curriculum design consists of two core components: difficulty measurer and training scheduler \cite{bengio2009curriculum}. Difficulty Measurer  is used to quantify the relative “easiness” of each data example. The training scheduler arranges the sequence of data subsets throughout the training process based on the judgment from the difficulty measurer. For ERC oriented curriculum learning, the challenge is how to design suitable difficulty measurer and training scheduler  for emotion recognition.


A conversation consists of a sequence of utterances. This hierarchical structure inspired us to construct two curricula for scheduling conversations and utterances respectively. Therefore, our framework consists of two nested curricula, conversation-level curriculum (CC) on the outside and utterance-level curriculum (UC) on the inside.

For CC, we  design an emotion-shift based difficulty measurer. A widely used CL strategy called \emph{baby step} \cite{spitkovsky2010baby}  is leveraged as training scheduler. 
For UC, due to the characteristics of the ERC task, the utterances in the same conversation must be input into a batch simultaneously during the training process. As a result, it is infeasible to employ traditional training scheduler such as \emph{baby step} to arrange the training order of the  utterances. We proposed emotion-similarity based curriculum learning to address this issue.

The proposed HCL framework is illustrated in Figure  \ref{fig:hcl} and the details of CC and UC are elaborated in following two subsections, respectively.

\subsection{Conversation-level Curriculum }

To design conversation-level curriculum for ERC, we need to answer a question: what kind of conversation is supposed to be easier than other conversations?
Since we have mentioned that previous ERC models  \cite{majumder2019dialoguernn,shen2021dialogxl} tend to suffer from emotion-shift issue, we adopt emotion-shift frequency to measure the difficulty of each conversation. The main idea is that, the more frequent emotion-shift in conversation $c_{i}$ occurs, the more difficult it is. Therefore, the conversation-level difficulty score of $c_{i}$ is defined as

\begin{equation}
\setlength{\abovedisplayskip}{0pt}
\setlength{\belowdisplayskip}{0pt}
d_{cc}(c_{i}) = \frac{N_{es}(c_{i}) + N_{sp}(c_{i})}{N_{u}(c_{i}) + N_{sp}(c_{i})}  ,   \label{equation:difficulty_score}
\end{equation}

where $N_{es}(c_{i})$ and $N_{u}(c_{i})$ denote the number of emotion-shift occurrences in $c_{i}$ and the total number of utterances in $c_{i}$, respectively. $N_{sp}(c_{i})$ is the number of speakers take part in $c_{i}$ and it acts as a smoothing factor.

We leverage baby step training scheduler  \cite{spitkovsky2010baby}  to arrange conversations and organize the training process. Specifically, the whole training set \emph{D} is divided into different buckets, i.e. $\{D_1, \cdots, D_T \}$, in which those conversations with similar difficulty scores are categorized into the same bucket. The training starts from the easiest bucket. After a fixed number of training epochs or convergence, the next bucket is merged into the current training subset. Finally, after all the buckets are merged and used, the whole training process further continues several extra epochs. Our HCL framework is described in Algorithm 1 and the process of CC is illustrated as \emph{Line 1-Line 5}.

\subsection{Utterance-level Curriculum }

As it is infeasible to employ a traditional  CL training scheduler  to asynchronously arrange the order of the utterances, the question arises how to measure the difficulty of the utterances and establish a feasible curriculum at utterance-level. 

We address this problem by assuming that the utterances with confusing emotion labels are more difficult for prediction and our utterance-level curriculum is based on the pairwise similarities between the emotion labels.

\begin{figure}[htbp]  	
	\centering  
	\includegraphics[scale=0.65]{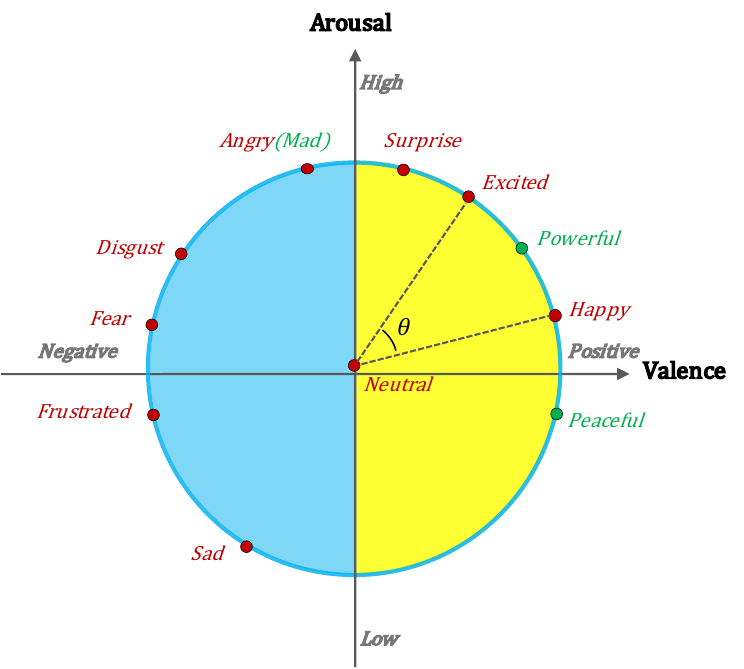}
	\caption{The 2D arousal-valence emotion wheel proposed by us. Each emotion label is corresponding to a point on the unit circle. This wheel has integrated the versions from \cite{jing2019automatic,yang2021circular,toisoul2021estimation}. The emotions in red color have appeared in previous versions. The emotions in green color is what we added (these emotions only appear in EmoryNLP dataset). $\theta$ denotes the intersection angle between \emph{happy} and \emph{excited}. The angle between similar emotions will be relatively small.  }  
	\label{fig:emotion_va}  	
\end{figure}

Previous studies  \cite{plutchik1982psychoevolutionary,mikels2005emotional,russell1980circumplex} on psychology believe that emotion contains two dimensions: arousal and valence, and they are used to leverage a wheel-like 2D coordinate system to describe emotions. Inspired by these works, we propose a new emotion wheel as Figure \ref{fig:emotion_va}, which contains all emotions in the standard ERC datasets. As depicted in Figure \ref{fig:emotion_va}, each emotion label can be mapped to a point on the unit circle. Then we calculate the similarity between emotion labels as in Equation \ref{eq:similarity}.

\begin{equation}
s_{ij}=\left\{
\begin{array}{lrc}
\max( \cos(\theta_{ij}), 0) & & {v_{i} \cdot v_{j} > 0}\\
0 & & {v_{i} \cdot v_{j} < 0}\\
1/N & & {v_{i} \cdot v_{j}  =  0}
\end{array} \right.
\label{eq:similarity}
\end{equation}

Here, $s_{ij}$ stands for the similarity of label $i$ and label $j$. $v_{i}$ denotes the valence value of $i$. We take the cosine of the included angle $\theta_{ij}$ between  $i$ and $j$  as their similarity. If $\theta_{ij} > 90^\circ$ (i.e., $cos\theta_{ij} < 0$) the similarity is set to 0. If the valence polarities of $i$ and $j$ are opposite, then the similarity is also set to 0.  The similarity between label \emph{neutral} and other labels is defined as $1/N$, where $N$ is the total number of emotions in corresponding datasets.


The process of emotion-similarity based curruclum learning (ESC) is described as \emph{Line 6 - Line 13} in Algorithm 1. We first calculate the similarity between each emotion label pair as Equation \ref{eq:similarity} and generate the emotion similarity matrix  $\mathcal{M}_{sim}$, then $\mathcal{M}_{sim}$ is normalized as $\mathcal{M}_{target}$.  At the beginning of ESC training, we take the rows of $\mathcal{M}_{target}$ as the initial target probability distributions over all possible classes for training, and each row corresponds to an emotion label. That is, instead of solely belonging to its ground-truth label, each input utterance can also belong to similar labels to a lesser extent. During the training process,  this label representation is gradually shifted towards the standard one-hot-encoding. We define the update strategy as in \emph{Line 9 - Line 11}, where $m_{i,j}$ denotes the probability of $j$-th element of $i$-th row in $\mathcal{M}_{target}$ at training step $t$. The constant parameter $\epsilon \in (0,1)$ controls how quickly the label vectors converge to the one-hot-encoded labels. Row-wise normalization is performed after each update. This update strategy leads to a proper label-weighting curriculum.

\begin{equation}
\setlength{\abovedisplayskip}{0pt}
\setlength{\belowdisplayskip}{0pt}
\mathcal{L}(\theta) = - \sum_{c=1}^{z} \sum_{i=1}^{n} \sum_{k=0}^{m} \mathcal{M}_{target}[y_{u_i}^c]_k \log\mathcal{P}_{u_i}^c[k]      \label{eq:loss}
\end{equation}

For each training step,  the predicted probability distribution of utterance  $u_i$ defined as $\mathcal{P}_{u_i}$. Finally, the model is trained with the standard cross-entropy loss function as Equation  \ref{eq:loss}, where $\mathcal{P}_{u_i}^c[k] $ denotes the predicted probability that the label of $u_i$ in conversation $c$ is $k$. $ \mathcal{M}_{target}[y_{u_i}^c]_k$ denotes the target probability of label $k$ in current label-similarity matrix at training step $t$. $z$ is total number of conversations in training set, $n$ is the utterance number of conversation $c$. In this way, we implement UC through ESC.

\begin{algorithm2e}[h]
	\caption{Training Process with HCL}  
	\LinesNumbered 
	\KwIn{
		\\$\mathcal{D}: \mbox{whole training dataset;}$ 
		\\$\mathcal{F}: \mbox{the difficulty measurer in CC;}$
		\\$k: \mbox{the number of buckets in baby step scheduler;}$ 
		\\$\mathcal{M}_{sim} : \mbox{the emotion similarity matrix in ESC}$ 
		\\$T: \mbox{the max training step for ESC;}$ 
		\\$\Delta{t}$: interval step for updating the targets in ERC;
		\\$\epsilon$: decay factor in ESC;
	}
	\KwOut{$M^*: \mbox{the optimal model.}$ }

	$ \mathcal{D}^{'} \ = \ \mbox{sort}(\mathcal{D}, \ \mathcal{F} )  $ \\ 
	$ \{ \mathcal{D}^1, \mathcal{D}^2, ... , \mathcal{D}^k \} = \mathcal{D}^{'} \ \mbox{where} \ \mathcal{F}(d_a) < \mathcal{F}(d_b), \ d_a \in \mathcal{D}^i, \  d_b \in \mathcal{D}^j,  {\forall i < j}
	$ \\
	$ \mathcal{D}^{train} = \emptyset  $ \\

	\For {$s = 1 \ ... \ k \ $} {
		$ \mathcal{D}^{train} \ = \ \mathcal{D}^{train} \cup  \mathcal{D}^s $ \\
		
		$\mathcal{M}_{sim} = \{s_{ij}\}, \ i, \ j=1,\cdots,m$ \\
		
		$ \mathcal{M}_{target} = \frac{ m_{ij}^{sim} }{ \sum_{j=1}^{m}m_{ij}^{sim}}, \ i,j=1,\cdots,m $ \\
		
		\For{$t = 1 \cdots T \ $}{
			\If {$  t \ \%  \ \Delta{t} = 0$} { 
				
				$m_{ij}^{'} =\left\{\begin{matrix} 
				\frac{1}{ 1+ \epsilon \sum_{j \neq j} m_{i,j}}, \ if \ , \ i = j \\
				\frac{\epsilon m_{ij}^t}{ 1+ \epsilon \sum_{j \neq j} m_{i,j}}, \ if \ , \ i \neq j
				\end{matrix}\right.$
				$ \mathcal{M}_{target} = \{ \frac{ m_{ij}^{'} }{ \sum_{j=1}^{m}m_{ij}^{'} } \}, \ i,j=1,\cdots,m $ \\ 
			}
			\mbox{train}($M$, $\mathcal{D}^{train}$, $\mathcal{M}_{target}$)
		}      
	}
	return M
\end{algorithm2e}

\section{Experimental Settings}

\subsection{Datasets}

We evaluate our method on the following four published ERC datasets \footnote{These datasets are multi-modal datasets, we only focus on the textual information so as to be consistent with previous works.}: \textbf{IEMOCAP} \cite{busso2008iemocap}, \textbf{MELD} \cite{poria2019meld}, \textbf{DailyDialog} \cite{li2017dailydialog}, \textbf{EmoryNLP} \cite{zahiri2018emotion}. The detailed statistics of the datasets are reported in Table \ref{tab:datasets_statics} \footnote{ Some baseline methods made slight adjustments in data splits, we keep exactly the same settings as corresponding methods respectively for fair comparison.}.

Following previous works  \citep{ghosal2019dialoguegcn,zhong2019knowledge,ishiwatari2020relation}, the evaluation metrics are chosen as micro-F1  excluding the extremely high majority class (neutral) for DailyDialog and weighted-F1 for other three datasets.

\begin{table*}
	\centering
	\footnotesize
	\begin{tabular}{c|c|c|c|c|c|c|c|c|c}
		
		\toprule
		
		\multirow{2}{*}{Datasets} & \multicolumn{3}{c|}{Conversations}  & \multicolumn{3}{c|}{Utterances} & \multirow{2}{*}{classes} & \multirow{2}{*}{avg\_utt} & \multirow{2}{*}{Evaluation} \\ 
		& Train & Val & Test & Train & Val & Test & & & \\
		
		\midrule
		IEMOCAP& \multicolumn{2}{c|}{120} & 31& \multicolumn{2}{c|}{5810} &1623& 6 &66.8& Weighted-F1\\ 
		
		MELD & 1038 & 114 & 280 & 9989 & 1109 & 2610 & 7 & 9.57 &Weighted-F1\\ 
		EmoryNLP & 713 & 99 & 85 & 9934 & 1344 & 1328 & 7 & 14.05 &Weighted-F1\\ 
		DailyDialog & 11118 & 1000 & 1000 & 87170 & 8069 & 7740 & 7 & 7.85 & Micro-F1\\
		
		\bottomrule
	\end{tabular}
	\caption{The statistics of datasets. $avg\_utt$ denotes the average number of utterances. } \label{tab:datasets_statics}
\end{table*}

\subsection{Baselines}

Since HCL is a model-agnostic framework, we choose the following five ERC models to verify whether HCL is able to further improve the performance of these models.

\paragraph{DialogueRNN}\cite{majumder2019dialoguernn}  This is a famous sequence-based ERC model, which uses three GRUs to model the speaker, the context given by the preceding utterances, and the emotion behind the preceding utterances. 

\paragraph{DialogueGCN} \cite{ghosal2019dialoguegcn} This is a representative graph-based ERC model. It captures self-dependency and inter-speaker dependency by using two-layer graph neural networks.

\paragraph{DAG-ERC} \cite{shen2021directed} It is the state-of-the-art of all the ERC models that do not employ external knowledge. DAG-ERC utilizes directed acyclic graph to model the structure of a conversation.

\paragraph{COSMIC} \cite{ghosal2020cosmic} It is a representative knowledge-based ERC model. It leverages external commonsense knowledge to improve the performance.

\paragraph{TODKAT} \cite{zhu2021topic} This is the state-of-the-art  knowledge-based ERC model. Besides commonsense knowledge, it also incorporates topic information.

\subsection{Implementation Details}

All of the baseline models mentioned above have released their source codes. We keep exactly the same settings as reported in the original papers during our experiments. For HCL, the tunable hyperparameters include number of buckets in CC, max training epochs during each baby step, interval steps for training target updating in UC, decay factor in UC. These hyperparameters are manually tuned on each dataset with hold-out validation. The results reported in our experiments are all based on the average score of 5 random runs on the test set. Our experiments are conducted on a single Tesla V100M32 GPU.

\section{Results and Analysis}

\subsection{Overall Results}

\begin{table*}
	\centering  
	\footnotesize 
	\begin{tabular}{cllll} 
		\toprule 
		METHOD&IEMOCAP&MELD&DailyDialog&EmoryNLP\\ 
		\midrule 	
		DialogueRNN&62.75&57.03&-&-\\ 
		DialogueGCN&64.18&58.10&-&-\\ 
		COSMIC&65.28&65.21&58.48&38.11\\ 
		DAG-ERC&68.03&63.65&59.33&39.02\\ 
		TODKAT&61.33&65.47&58.47&43.12\\ 
		\midrule 	
		DialogueRNN+HCL&64.62 $(\uparrow{1.87})$ & 58.30 $(\uparrow{1.27})$&-&-\\ 
		DialogueGCN+HCL&65.41 $(\uparrow{1.23})$ & 59.31 $(\uparrow{1.21})$&-&-\\ 
		COSMIC+HCL&66.23$(\uparrow{0.95})$&65.85$(\uparrow{0.64})$&59.54$(\uparrow{1.06})$&38.96$(\uparrow{0.85})$\\
		DAG-ERC+HCL&\bf{68.73}$(\uparrow{0.70})$  &63.89$(\uparrow{0.24})$&59.64$(\uparrow{0.31})$&39.82$(\uparrow{0.80})$\\ 
		TODKAT+HCL&63.03$(\uparrow{1.70})$&\bf{66.18}$(\uparrow{0.71})$&\bf{59.76}$(\uparrow{1.29})$&\bf{46.11}$(\uparrow{2.99})$\\ 
		
		\bottomrule 
	\end{tabular}  
	\caption{ The overall results on different methods on four datasets. The results of baseline methods are from the original papers.} \label{tab:performance_comparison}  
\end{table*}

The overall experimental results are reported in Table \ref{tab:performance_comparison}, where ``X+HCL'' means training the model X with the proposed HCL framework. We can see that HCL has improved the performance of all baseline models, showing the robustness and universality of our approach. 

In general, the performance boosts achieved by HCL on models with simpler feature extractor (i.e., DialogueRNN and DialogueGCN) are more remarkable. An exception is that TODKAT+HCL achieves significant improvements on three datasets. The reason may be that the original TODKAT model does not take account of the speaker information, while our CC has introduced the inter-speaker emotion-shift  in the difficulty measurer, which is equivalent to considering speaker information to a certain extent and is beneficial for TODKAT.

\subsection{Ablation Study}

\begin{table}[h]
	\centering  
	\footnotesize 
	\begin{tabular}{cll} 
		\toprule 
		METHOD&IEMOCAP&EmoryNLP\\ 
		\midrule   
		TODKAT&61.33&43.12\\
		TODKAT+CC&61.83$(\uparrow{0.50})$&44.20$(\uparrow{1.08})$\\ 
		TODKAT+UC&62.01$(\uparrow{0.68})$&45.19$(\uparrow{2.07})$\\ 
		\midrule   
		TODKAT+CCF&62.07$(\uparrow{0.74})$&45.06$(\uparrow{1.94})$\\ 
		TODKAT+UCF&62.76$(\uparrow{1.43})$&45.47$(\uparrow{2.35})$\\     
		TODKAT+HCL&\bf{63.03}$(\uparrow{1.70})$&\bf{46.11}$(\uparrow{2.99})$\\ 
		\bottomrule 
	\end{tabular}  
	\caption{ Ablation study on TODKAT} \label{tab:ablation_studies}  
\end{table}

To reveal the individual effects of CC and UC, we try different variants of HCL on TODKAT by removing either CC or UC. The experimental results on IEMOCAP and EmoryNLP are shown in Table \ref{tab:ablation_studies}, from which we see that both CC and UC make positive contributions to the overall performance when used alone. Although only utilizing UC leads to larger improvements than only using CC, the optimal performance is achieved when CC and UC are combined, indicating that CC and UC are complementary to each other. 

In addition, we also tried another two strategies to combine CC and UC: CC-First (CCF) and UC-First (UCF). CCF performs CC and UC in a pipeline manner. In UCF, the execute order of CC and UC is reversed. The results of CCF and UCF are also outlined in Table \ref{tab:ablation_studies}. It shows that UCF is better than CCF and HCL  outperforms both CCF and UCF. This is intuitive, because  HCL makes UC and CC interact with each other during the training process, which is more consist with the hierarchical structure of conversation, so the performance is even better than UCF.

\subsection{Performance for Emotion-shift}

\begin{table}[h]
	\centering
	\footnotesize
	\begin{tabular}{c|c|c|c|c}
		
		\toprule
		
		\multirow{2}{*}{METHOD} & \multicolumn{2}{c|}{IEMOCAP}  & \multicolumn{2}{c}{EmoryNLP}\\ 
		& ES & N-ES &  ES & N-ES\\
		& (41.2\%) & (58.8\%) & (69.2\%) & (30.8\%) \\
		\midrule
		TODKAT & 56.24 & 64.62 & 39.36 & 51.51\\
		TODKAT+HCL & \bf{56.91} & \bf{67.01} & \bf{42.40} & \bf{54.02}\\ 
		
		\bottomrule
	\end{tabular}
	\caption{The performance of TODKAT+HCL on utterances which exhibits emotion-shift. Numbers in parenthesis indicate the percentage in the test dataset.} \label{tab:es}  
\end{table}

To verify the effect of HCL in the emotion-shift scenario, we summarize the results of TODKAT+HCL on different types of utterances. The results are presented in Table \ref{tab:es},  where \emph{ES} and  \emph{N-ES} denote utterances with emotion-shift and utterances without emotion-shift, respectively. HCL improves the performance of  TODKAT on both  \emph{ES}  and  \emph{N-ES} of the two datasets. The improvement on ES in EmoryNLP is more significant than on ES in IEMOCAP. 

A plausible explanation is that the training set of IEMOCAP contains much less conversations and the average length of conversations is much longer, so the difficulty scores of conversations in IEMOCAP are usually lower. Therefore, for IEMOCAP, the difficulty discrimination between different buckets in the training scheduler is not as obvious as EmoryNLP.

 \subsection{Performance  on Different Emotions}

 In this subsection, we aim to verify whether HCL can improve the  performance of baseline model on ``confusing labels''. For each pair of emotion labels in ERC dataset, if their similarity (defined in Equation \ref{eq:similarity}) is larger than 0, then both of them are regarded as  confusing labels in our setting. \footnote{  \emph{Neutral} is not included}. We report the results of DAG-ERC and DAG-ERC+HCL on every emotion label in IEMOCAP. There are a total of four confusing labels in this dataset: \emph{happy}(H), \emph{excited}(E), \emph{sad}(S) and \emph{frustrated}(F). As presented in Table  \ref{tab:similarlabels}, DAG-ERC+HCL outperforms DAG-ERC on all emotion labels other than \emph{neutral} and the overall performance on the confusing labels is better ( \emph{69.37} vs \emph{67.88} on weighted-F1). This shows that HCL does strengthen the  ability on distinguishing the confusing emotion labels of DAG-ERC. However, the performace is limited by \emph{neutral}, the reason is that  \emph{neutral} is similar to every other label to some extent as in Equation \ref{eq:similarity}, which increases the difficulty for recognition.

 \begin{figure*}[h]  	
	\centering  
	\includegraphics[scale=0.70]{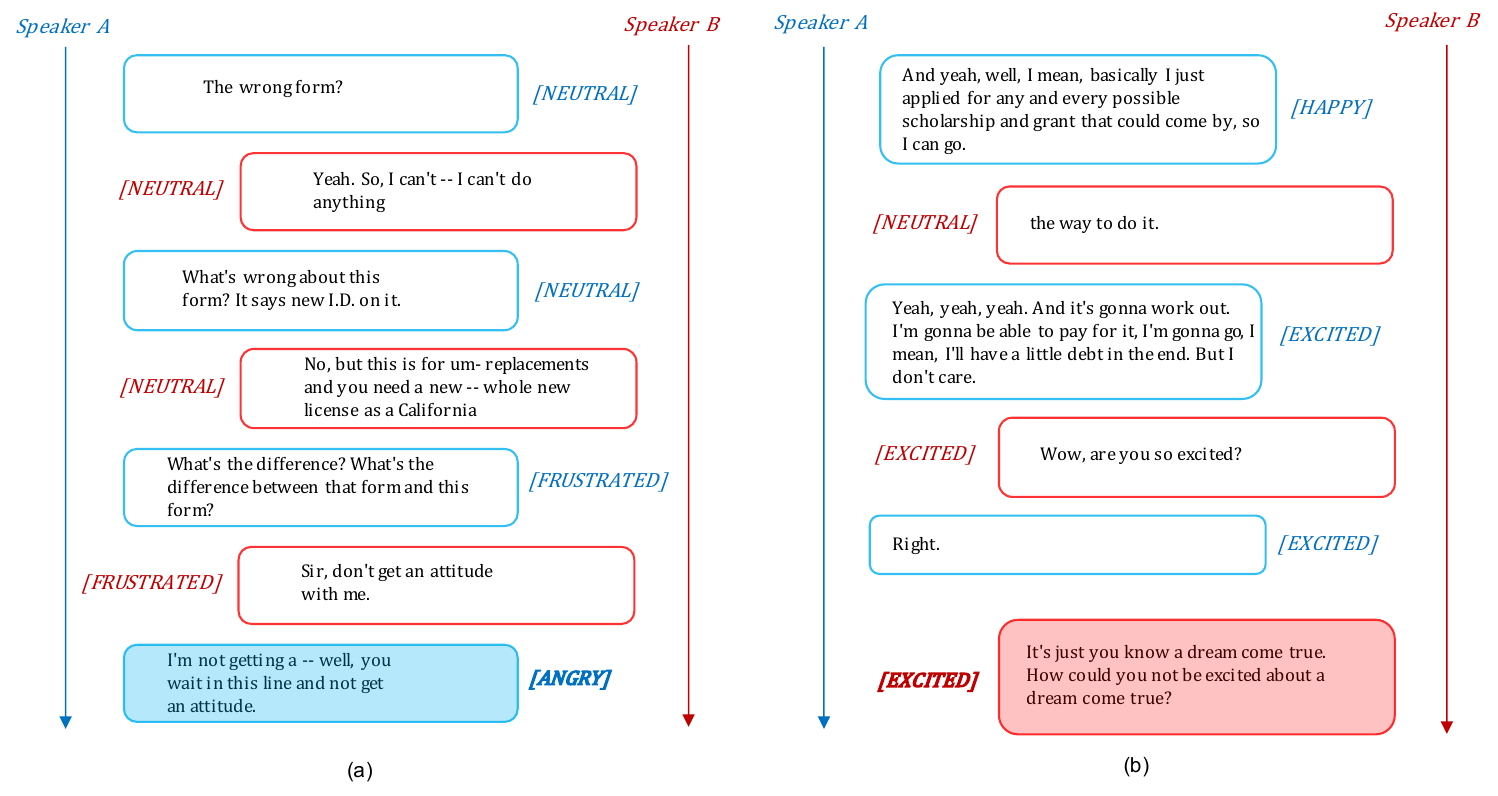}
	\caption{Two conversation passages from IEMOCAP for case study. The ground-truth emotion label of each utterance is given in the corresponding position. (a) An  emotion-shift case. (b) A confusing label case.}  
	\label{fig:case_study}  	
\end{figure*}

\begin{table*}
	\footnotesize
	\centering
	\begin{tabular}{ccccccccc}
		
		\toprule
		\multirow{2}{*}{Method}& Happy & Sad & Neutral & Angry& Excited & Frustrated & HESF & NA \\ 
		
		&(8.8\%)& (15.1\%) & (23.7\%) & (10.5\%) & (18.4\%) &(23.5\%)  & (65.8\%) & (34.2\%)\\
		
		\midrule 
		DAG-ERC & 47.59 & 79.83 & \bf{69.36} & 66.67 & 66.79	 & 68.66  & 67.88 & \bf{68.53}  \\
		DAG-ERC+HCL & \bf{48.97} & \bf{82.21} & 68.08 &	\bf{66.72} & \bf{69.43} & \bf{68.73}  & \bf{69.37} & 67.66  \\
		\bottomrule  
	\end{tabular}
	\caption{ Comparison of DAG-ERC and DAG-ERC+HCL on different emotions. Here \emph{HESF}  and \emph{NA} denote ``\emph{Happy + Excited + Sad + Frustrated}''  and ``\emph{Neutral + Angry}'', respectively. Numbers in parenthesis indicate the percentage of each emotion label in the test dataset.} \label{tab:similarlabels}
\end{table*}

\subsection{Case Study}

Figure \ref{fig:case_study}(a) shows a conversation passage sampled from the IEMOCAP dataset. The goal is to predict the emotion label of the last utterance in the blue box. Due to emotion-shift occurs, all the baseline methods in our experiment are easy to mistakenly identify the emotion as \emph{frustrated}. Most of our ``X+HCL'' methods are able to recognize the emotion of this utterance correctly, which indicates that HCL alleviates this problem to some extent.
Figure  \ref{fig:case_study}(b)  depicts a case with confusing labels. The gold emotion label of the last utterance in the red box is \emph{excited}. Some of the baseline models such as DialogueGCN and DAG-ERC mistook the emotion as \emph{happy}. After following HCL framework, DialogueGCN+HCL and DAG-ERC+HCL successfully identified the emotion as the correct label  \emph{excited}.

\subsection{Why Curriculum Learning Works?}

According to the theory of curriculum learning \cite{bengio2009curriculum}, the curriculum will work only if the entropy of data distributions increases during the training process. In HCL, conversation-level curriculum leverages the emotion-shift frequency to measure the difficulty. The more frequent the emotion-shift occurs in a conversation, the greater the diversity of the emotion labels, in other words, the higher the entropy. For utterance-level curriculum, since emotion-similarity based CL does not distinguish similar emotion in the early stage, it is equivalent to merging some emotion labels and could be considered as reducing the diversity of emotions. As a result, it also meets the condition which the entropy should be increased gradually.

\section{Conclusions}

In this paper, we propose simple but effective hybrid curriculum learning (HCL) for emotion recognition in conversations. HCL is a flexible framework independent of the original training models. During training, HCL simultaneously employs conversation-level and utterance-level curricula to execute the training process as an easy to hard schema. Conversation-level curriculum consists of an emotion-shfit based difficulty measurer and a baby step scheduler. Utterance-level curriculum is implemented as emotion-similarity based CL. Experiments on four benchmark datasets have proved the generality and effectiveness of HCL. 

In the future, we plan to improve our method in three directions. First, we will attempt to seek other suitable features to construct difficulty measurer for ERC. Second, we aim to introduce other training schedulers for CL to further improve the performance. Finally, we aim to apply a learning-based approach to model the similarity between emotion labels.

\bibliography{aaai22.bib}

\end{document}